\relax
\documentclass[letterpaper]{article} 
\usepackage{aaai19_arxiv}  
\usepackage{times}  
\usepackage{helvet}  
\usepackage{courier}  
\usepackage{url}  
\usepackage{graphicx}  
\usepackage[usenames, dvipsnames]{color}
\usepackage{amsmath}
\usepackage{svg}
\usepackage[warn]{textcomp}
\usepackage[inline]{enumitem}
\usepackage{booktabs}       

\makeatletter
\newcommand{\printfnsymbol}[1]{%
  \textsuperscript{\@fnsymbol{#1}}%
}
\makeatother

\begin{document}
\title{CAPNet: Continuous Approximation Projection For 3D Point Cloud Reconstruction Using 2D Supervision}
\author{Navaneet K L\thanks{equal contribution},
Priyanka Mandikal\printfnsymbol{1},
Mayank Agarwal, \and
R. Venkatesh Babu \\
Video Analytics Lab, CDS, Indian Institute of Science, Bangalore, India \\
navaneetl@iisc.ac.in, priyanka.mandikal@gmail.com, mayankgrwl97@gmail.com, venky@iisc.ac.in
}
\maketitle

\begin{abstract}
    
    Knowledge of 3D properties of objects is a necessity in order to build effective computer vision systems. However, lack of large scale 3D datasets can be a major constraint for data-driven approaches in learning such properties. We consider the task of single image 3D point cloud reconstruction, and aim to utilize multiple foreground masks as our supervisory data to alleviate the need for large scale 3D datasets.
    A novel differentiable projection module, called \lq{}CAPNet\rq{}, is introduced to obtain such 2D masks from a predicted 3D point cloud. The key idea is to model the projections as a continuous approximation of the points in the point cloud. To overcome the challenges of sparse projection maps, we propose a loss formulation termed  \lq{}affinity loss\rq{} to generate outlier-free reconstructions. We 
    significantly  outperform the existing projection based approaches on a large-scale synthetic  dataset. We show the utility and generalizability of such a 2D supervised approach through experiments on a real-world dataset, where lack of 3D data can be a serious concern. To further enhance the reconstructions, we also propose a test stage optimization procedure to obtain reconstructions that display high correspondence with the observed input image.
\end{abstract}
\section{Introduction}
\label{intro}

3D Reconstruction from images is a key challenge in the field of computer vision. 
While deep learning based approaches have achieved exceptional results in various computer 
vision tasks~\cite{krizhevsky2012imagenet,girshick2014rich,long2015fully,goodfellow2014generative}, the capability of such approaches is limited by the amount of data 
available. Obtaining large scale 3D data of objects can be expensive and 
time-consuming. In contrast, capturing 2D data (image, foreground mask etc.) from multiple 
view points is relatively easy. 
We consider the task of single image 3D point cloud reconstruction and aim to utilize such 2D observations in place of point clouds as our supervisory data. Towards this end, we propose a novel differentiable projection module to obtain the 2D observations from the predicted points. 

The nature of the projection module is dependent on our choice of the 3D representation. Unlike 2D images, where all the pixels add rich spatial and structural information, 
volumetric representations suffer from sparsity of information. The information needed to perceive the 3D structure is provided by surface voxels, 
while the voxels within the volume increase the representational and computational 
complexity with minimal addition in 
information. 3D point clouds are a more efficient alternative, since the points are sampled on the surface of the object. 
However, lack of grid structure and permutation invariance 
properties of point clouds pose challenges in their processing. Recent works 
address these issues using point cloud specific architectures and loss formulations 
~\cite{qi2017pointnet,qi2017pointnet++,fan2017point,su2018splatnet,li2018pointcnn}.

In the case of 3D voxel-based approaches, the projection is obtained via a transformation between two grid representations and hence can be performed by simple interpolation operations~\cite{yan2016perspective}. Point clouds, however, pose two important challenges: 
\begin{enumerate*}[label=\textbf{(\arabic*)}]
    \item Firstly, projecting low density point clouds using conventional interpolation techniques can result in projections with holes. Generating high density point clouds requires higher memory and computational power.
    \item Secondly, the process of obtaining the projections by discretizing the point cloud is a non-differentiable operation.
\end{enumerate*}
To address both these issues, we propose a continuous approximation of points in the point cloud which produces smooth projections in a differentiable manner. 
However, a sparse projection map results in very low gradients in regions where no point is projected, which leads to outlier points. We propose a novel loss function, termed \textit{Affinity Loss}, to enforce effective flow of gradients in such situations, resulting in cleaner and better reconstructions.      

Since 2D observations like foreground masks can be obtained from the input image, 
projection based approaches provide a unique opportunity for optimization on the test
data. Given a test image, the point cloud obtained using a pre-trained networked can
be modified to exactly match the corresponding mask. We introduce such a set-up and 
obtain more accurate reconstructions with improved correspondence to the input image.

To summarize, we make the following key contributions:
\begin{itemize}
    \item We propose CAPNet, a continuous approximation projection module for a differentiable and accurate rendering of 3D point clouds, to enable weakly supervised 3D object reconstruction. The proposed rendering module generates smooth, artifact-free projections, while also overcoming the lack of gradients that can exist in a naive discretization based approach.
    \item We formulate a loss function termed \textit{Affinity Loss} for effectively penalizing outlier points, resulting in reconstructions of high quality.
    \item Using as little as a single mask as supervision, we fine-tune a 3D supervised network on a real world dataset, and demonstrate the efficacy of the approach in obtaining superior reconstructions.
    \item We perform extensive quantitative and qualitative evaluation of CAPNet on synthetic and real datasets, and show that it significantly outperforms the state-of-the-art projection-based reconstruction methods.
    \item Finally, we present a technique for optimizing the predicted point clouds at test time using available foreground masks, and obtain reconstructions that highly correspond to the input image. 
\end{itemize}    


\section{Related works}

\noindent
\textbf{3D Supervision}

A number of 3D reconstruction works employ training procedures that utilize the complete 3D data available. With the advent of deep neural network architectures in 2D image generation tasks, the power of convolutional neural nets have been directly transferred to the 3D domain using 3D CNNs. There is vast literature on generating voxelized output representations.
~\cite{girdhar2016learning} learnt a joint embedding of 3D voxel shapes and their corresponding 2D images.~\cite{choy20163d} trained a recurrent neural network to encode information from many input views. These works predict voxelized outputs and utilize 3D voxel-based losses for supervision. But the compute overhead and sparsity of information in voxel formats inspired lines of work that abstracted volumetric information into smaller number of units with the help of the octree data structure~\cite{tatarchenko2017octree,Riegler2017CVPR,hspHane17}. More recently, Fan \textit{et al.}~\cite{fan2017point}, introduced frameworks and loss formulations tailored for generating unordered point clouds, and achieved single-view 3D reconstruction results outperforming the volumetric state-of-art approaches~\cite{choy20163d}. Several other recent works tackle the problem of 3D point cloud reconstruction from a single image~\cite{groueix2018,mandikal20183dlmnet,mandikal20183dpsrnet,mandikal2019densepcr}. While all of the above works directly use full 3D point cloud supervision, we show competitive 3D reconstruction capability by using only 2D masks as supervision.

\vspace{0.5em}
\noindent
\textbf{2D Supervision}

Recent works have explored ways to reconstruct 3D shapes from 2D projections such as silhouettes and depth maps with the help of differentiable rendering modules.
\begin{enumerate}[label=\textbf{(\arabic*)}]
    \item \textbf{Volumetric Rendering}: Perspective transformer nets (PTN)~\cite{yan2016perspective} performs perspective transformation and grid sampling of volumetric outputs to obtain the projections. Tulsiani \textit{et al.}~\cite{tulsiani2017multi} use differentiable ray consistency loss to train on 2D observations like foreground mask, depth, color images and semantic maps. MarrNet~\cite{wu2017marrnet} predicts normal, depth and silhouette maps from 2D images and reconstructs voxel outputs using the estimated 2.5D sketches, while also enforcing projection consistency at test time.

    \item \textbf{Point-cloud Rendering}: Amongst the point-cloud based works, Lin \textit{et al.}~\cite{lin2018learning} use a combination of depth fusion and point projection as supervision. Our approach differs from Lin \textit{et al.} in the following ways:
    \begin{enumerate*}[label=\textbf{(\alph*)}]
        \item In Lin \textit{et al.}, the network is first pretrained without using projection loss, but directly regressing for depth maps from eight fixed views, which are fused to obtain the point cloud. The projection loss is only used for fine-tuning this model once it has been trained. On the other hand, our model requires no pretraining and as little as a single mask from a random view can be used as supervision.
        \item The projection module in Lin \textit{et al.} consists of discretizing the $(x,y)$ coordinates of the point cloud and projecting the $z$ values onto a plane. To avoid point collisions, a memory intensive operation (termed 'upsampling') is performed. In contrast, we directly obtain the projected map via the continuous approximation module. Further, we introduce the \textit{Affinity Loss} in the training regime to remove outlier points.
\end{enumerate*}

\end{enumerate}

Apart from volumetric and point-cloud based approaches, differentiable rendering modules for 3D meshes have also been proposed~\cite{kato2017neural}.

\section{Approach}
\subsection{Problem Formulation}
\label{sec:prob_formulation}
Given a single image of an object, we aim to reconstruct its 3D point cloud 
representation. Let $I$ be an image
from the training set. Let $p=f(I)$ be the corresponding 3D point cloud
reconstruction obtained using the network $f(.)$. A projection $P(p,v)$ from an arbitrary view point $v$ is
obtained by performing a perspective transformation and projecting the
transformed point cloud on to a plane. The view-point is parametrized by
the camera calibration matrix and extrinsic parameters $K$ and $(R,t)$ respectively. We
assume the knowledge of these parameters in the training stage. Let $N$ be the 
number of points in the point cloud. Then the transformed
point $\hat{p}_n=(\hat{x}_n,\hat{y}_n,\hat{z}_n)$ in the camera coordinates is obtained as follows: 
\begin{equation}
    \hat{p}_n = K(R_{v}p_n+t_{v}) \quad \forall n \in \{1,\cdot\cdot\cdot,N\}    
\end{equation}
To train the 3D reconstruction network, the ground truth 2D mask, $M$ is used to supervise the projection, $\hat{M}=P(p,v)$. 
\begin{figure*}[!htb]
\centering
\begin{center}
    \includegraphics[width=\linewidth]{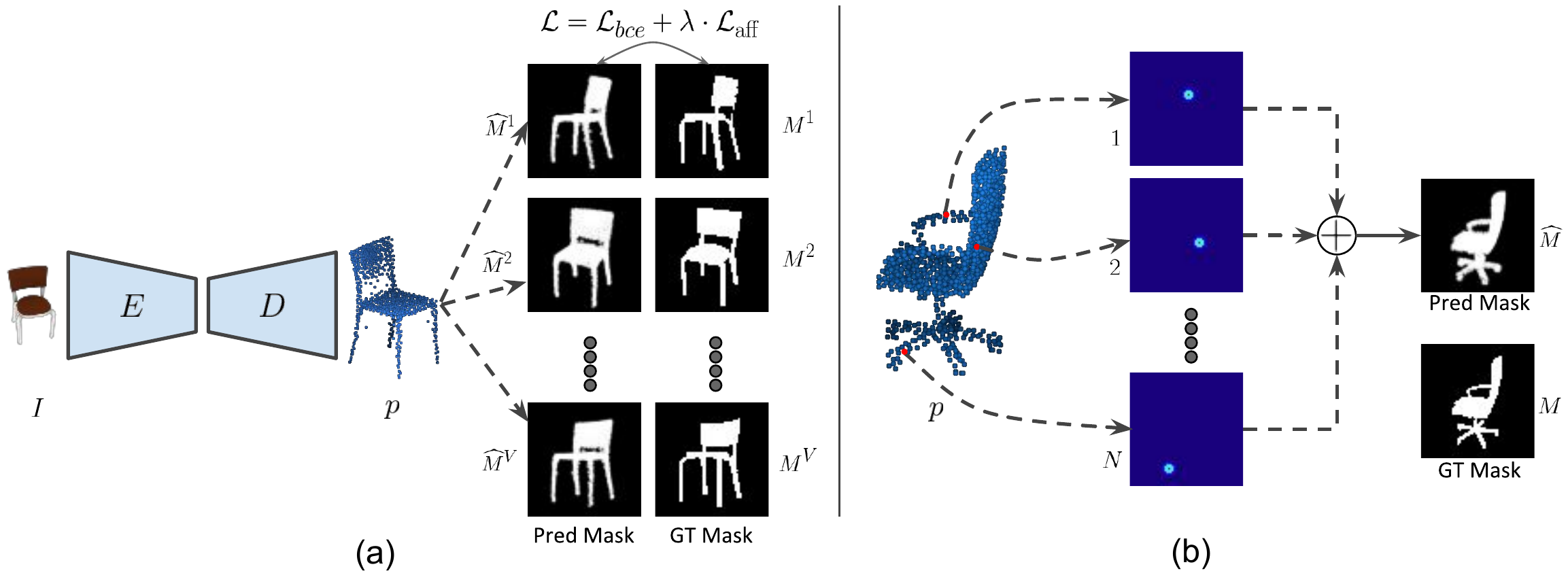}
\end{center}
\caption{\textbf{Network architecture and projection module}: (a) An encoder-decoder 
architecture is used to obtain point cloud reconstructions from a 2D image. The 
point cloud is projected from multiple view points and compared with corresponding 
ground truth mask. We use a combination of binary cross-entropy and point affinity loss as 
our projection consistency loss. (b) An overview of our projection module is shown. 
Each point in the prediction (red dot in the image) is projected on to a 2D grid by generating a Gaussian map 
centered at the (x,y) location of the the point. The Gaussian maps from all the points
are then combined to obtain a smooth projection that matches the ground truth. Image best viewed zoomed and in color.}
\label{fig:network_arch}
\end{figure*}
\subsection{Continuous Approximation Projection} 
The 3D reconstruction network consists of an encoder 
which takes in a 2D image as input, followed by a decoder which reconstructs the point cloud (Fig.~\ref{fig:network_arch}). The predicted points 
are projected from $V$ different view-points and the loss is calculated with the corresponding ground truth projections.
Let $\hat{M}_{i,j}^v$ be the pixel value of $v^{th}$ projected mask at $(i,j)$ coordinates. 
The projected mask is obtained as follows: 
\begin{equation}
    \hat{M}_{i,j}^{v} = tanh\Bigg(\sum \limits_{n=1}^{N} \phi(\hat{x}_{n}-i) \phi(\hat{y}_{n}-j)\Bigg)
\end{equation}
where $tanh$ is the hyperbolic tangent function and $\phi(.)$ is a 
kernel function. 
To obtain a smooth projection, we use an un-normalized Gaussian kernel of variance $\sigma^2$:
\begin{equation}
    \phi(k) = \text{exp}\Big(\frac{-k^{2}}{2\sigma^{2}}\Big)
    \label{eq:kernel}
\end{equation}
The variance of the kernel is set such that the projection is smooth while retaining 
the finer structures present in the ground truth mask. Refer to Discussion Section for details.

The proposed point cloud rendering module is significantly different from the existing volumetric approaches for the following reasons:
\begin{enumerate*}[label=\textbf{(\arabic*)}]
    \item Unlike an ordered voxel representation, a point cloud does not reside in a discretized grid world, but rather in continuous free space. A direct application of volumetric rendering would require embedding the point cloud in a 3D grid, and such a discretization operation is non-differentiable, preventing back-propagation in neural networks. We navigate this problem by treating the value at every pixel to be a continuous approximation of the points in the point cloud.
    \item Volumetric rendering in~\cite{yan2016perspective} is handled by the Spatial Transformer Network (STN)~\cite{jaderberg2015spatial}, which performs bilinear interpolation at the grid cell corners. Apart from being non-differentiable for point sets, this approach would produce 'holes' in the projection for low density point clouds. On the contrary, we introduce a continuous approximation module which utilizes a Gaussian kernel to obtain smooth and accurate projections (Discussion Section, Fig.~\ref{fig:projection_maps}b).
\end{enumerate*}

\subsection{Loss Formulation}
We enforce consistency between the projected and ground truth maps using the binary cross-entropy loss, given by:
\begin{equation}
    \mathcal{L}_{bce}  = \sum \limits_{v=1}^{V} -M^{v} \text{log}\hat{M}^{v} - (1-M^{v}) \text{log}(1-\hat{M}^{v})
    \label{eq:loss_bce}
\end{equation}
where $M^{v}$ and $\hat{M}^{v}$ are the ground truth and predicted masks respectively of dimension $(H,W)$.
However, we observe that training the network with just $\mathcal{L}_{bce}$ results in reconstructions with a large number of outlier points. To alleviate this effect, we propose a loss function that penalizes outlier points in the projected maps by enforcing a \textit{nearest point affinity loss}, defined as follows:
\begin{subequations}
    \begin{align}
        \begin{split}
            \mathcal{L}_{\mathrm{aff}} &= \sum\limits_{v=1}^{V} \sum\limits_{i,j}^{H,W} \min\limits_{(k,l)\in M_{+}^{v}} ((i-k)^2 + (j-l)^2)\hat{M}_{i,j}^{v} M_{k,l}^{v} \\ 
            & + \sum\limits_{v=1}^{V} \sum\limits_{i,j}^{H,W} \min\limits_{(k,l)\in \hat{M}_{+}^{v}} ((i-k)^2 + (j-l)^2)M_{i,j}^{v}\hat{M}_{k,l}^{v}
        \end{split}
    \end{align}
    \label{eq:loss_aff}
\end{subequations}
where $M_+^{v}$ and $\hat{M}_+^{v}$ are sets of pixel coordinates of the ground truth and predicted projections whose values are non-zero. Intuitively, this constraint minimizes the nearest neighbour distance between two pixel maps weighted by pixel confidence.
We observe that the use of $\mathcal{L}_{\mathrm{aff}}$ is critical in obtaining meaningful reconstructions (Discussion Section). Regions in the mask where ground truth confidence is one, but the prediction is near zero, might result in weak gradients if there are no predicted points in the nearby region. Similar issues arise when an incorrect isolated prediction is present. Affinity loss helps in alleviating both these issues.
The final loss function during optimization is a combination of binary cross-entropy and affinity loss:
\begin{equation}
    \mathcal{L} = \mathcal{L}_{bce} + \lambda \cdot \mathcal{L}_{\mathrm{aff}}
    \label{eq:loss}
\end{equation}

\subsection{Test-Stage Optimization (TSO)}
While the reconstructions obtained by training a network as described above are reasonably correct, they fail to exactly match the input image. Existing state-of-the-art approaches also produce outputs that do not correspond completely, and often fail to reconstruct finer details visible in the image. We propose a test-stage optimization procedure that starts from an initial point cloud $y$, that is the output from any point cloud prediction network, and outputs an optimized point cloud $\hat{y}$. This procedure aims to exactly match the reconstructed point cloud in the region visible in the input image, while obtaining meaningful structure in the occluded regions. To achieve this, we match the projection of the point cloud from the input image view. We make the reasonable assumption of the presence of ground truth mask for the input image in the test stage. We explore three ways of updating the predicted point cloud:
\begin{enumerate}[label=\textbf{(\arabic*)}]
    \item \textbf{Update only $E$}: Keeping the decoder fixed, we only update the parameters of the encoder by optimizing for the binary cross entropy loss $\mathcal{L}_{bce}$ (Eqn.~\ref{eq:loss_bce}).
    \item \textbf{Update $E$ and $D$}: We update the parameters of both the encoder and decoder to match the projection. In order to preserve the structure in the occluded regions, we additionally employ a regularization loss on the update of the point cloud. The total loss is defined as
    \begin{equation}
    \label{eqn:tso}
    \mathcal{L}_{tso} = \mathcal{L}_{bce} + \gamma \cdot d(y,\hat{y})    
    \end{equation}
    where $d(.)$ is a distance function, $y$ and $\hat{y}$ are the initial and optimized point clouds respectively.
    \item \textbf{Directly update $p$}: We directly update the predicted points using Eqn.~\ref{eqn:tso}, without updating any of the network parameters. This approach enables the optimization procedure to be utilized even in cases where the point cloud prediction network is not available during inference.
\end{enumerate}

\section{Experiments}
\label{sec:experiments}
\subsection{3D Reconstruction on ShapeNet}

\paragraph{Implementation details} We fix the number of projections to four in all the experiments. The view-points are randomly selected as in \cite{tulsiani2017multi}. 
The kernel variance $\sigma^2$ in Eqn.~\ref{eq:kernel} is chosen as [$0.4,0.4,0.1$] for [chair,car,airplane] in the single-category experiments and $0.4$ in the multi-category experiments. $\lambda$ is set to $1$ in Eqn.~\ref{eq:loss}. We use Adam optimizer to train the network with a learning rate of $5e^{-5}$. The network architecture details are provided in the supplementary material. For the test-stage optimization procedure, we experiment with different hyperparameter settings for each of the three variants and choose settings that are optimal for each. Learning rates are set to $1e^{-6}$, $5e^{-6}$, and $5e^{-4}$ for variants 1, 2 and 3 respectively. The weightage for regularization is set to $1e^{6}$ in Eqn.~\ref{eqn:tso}. The optimization procedure is run for 50 iterations and takes 1s on an Nvidia GTX 1080Ti GPU.
\paragraph{Dataset} We use the textured 3D CAD models from the synthetic ShapeNet dataset~\cite{chang2015shapenet} for our experiments. We consider three exemplar categories: airplane, car and chair. We follow the set-up of  
~\cite{tulsiani2017multi} and use the same train/val/test splits so as to be comparable to existing works. 
\paragraph{Evaluation Methodology} We use the Chamfer distance between point clouds as our metric to evaluate reconstruction. The Chamfer distance between two point clouds
$\widehat{P}$ and $P$ is defined as $d_{Chamfer}(\widehat{P},P) = \sum_{x\in \widehat{P}}\min_{y\in P}{||x-y||}^2_2 + \sum_{x\in \widehat{P}}\min_{y\in P}{||x-y||}^2_2$.
The ground truth point cloud is obtained by randomly sampling 16,384 points on the surface of the object and performing farthest point sampling to obtain 1024 points. To evaluate approaches which reconstruct voxelized representations, we use the code provided by~\cite{pix3d} to convert them to point clouds. The procedure consists of first generating an iso-surface from the voxels using the Marching Cubes algorithm ~\cite{lorensen1987marching}, and then sampling points on the generated surface to obtain the final point cloud. 
\begin{table*}[!htb]
  \caption{Chamfer metrics on ShapeNet (all values are scaled by 1000). We significantly outperform both mask and depth variants of DRC, and obtain scores close to 3D supervised PSGN-FC. In comparison to Lin et al., who fuse depth maps from eight fixed views, we obtain better performance using just mask projection from four random views.}
  \centering
  \begin{tabular}{llllll}
    \toprule
    Method              & Supervision    & Airplane  & Car      & Chair  & Mean   \\
    \midrule  
    PSGN-FC~\cite{fan2017point}             & 3D     & 1.36      & 1.40     & 3.98   & 2.25   \\
    PSGN-FC (multi-cat)~\cite{fan2017point} & 3D     & 1.33      & 1.41     & 3.94   & 2.23   \\
    \midrule  
    DRC-Depth~\cite{tulsiani2017multi}   & Depth  & 6.30      & 4.33     & 11.38   & 7.34   \\
    Lin et al.~\cite{lin2018learning}   & Depth     & 2.01  & 2.50     & 6.35   & 3.62   \\
    \midrule  
    DRC-Mask~\cite{tulsiani2017multi}    & Mask   & 18.94     & 4.92     & 15.91   & 13.26   \\
    Ours           & Mask           & \textbf{2.00}       & \textbf{1.65}     & \textbf{4.42}   & \textbf{2.69}   \\
    Ours (multi-cat)    & Mask      & 2.57      & 1.74     & 5.09  & 3.13   \\
    \bottomrule
  \end{tabular}
  \label{tab:shapenet_mask}
\end{table*}
\begin{figure*}[!htb]
\centering
\begin{center}
    \includegraphics[width=\linewidth]{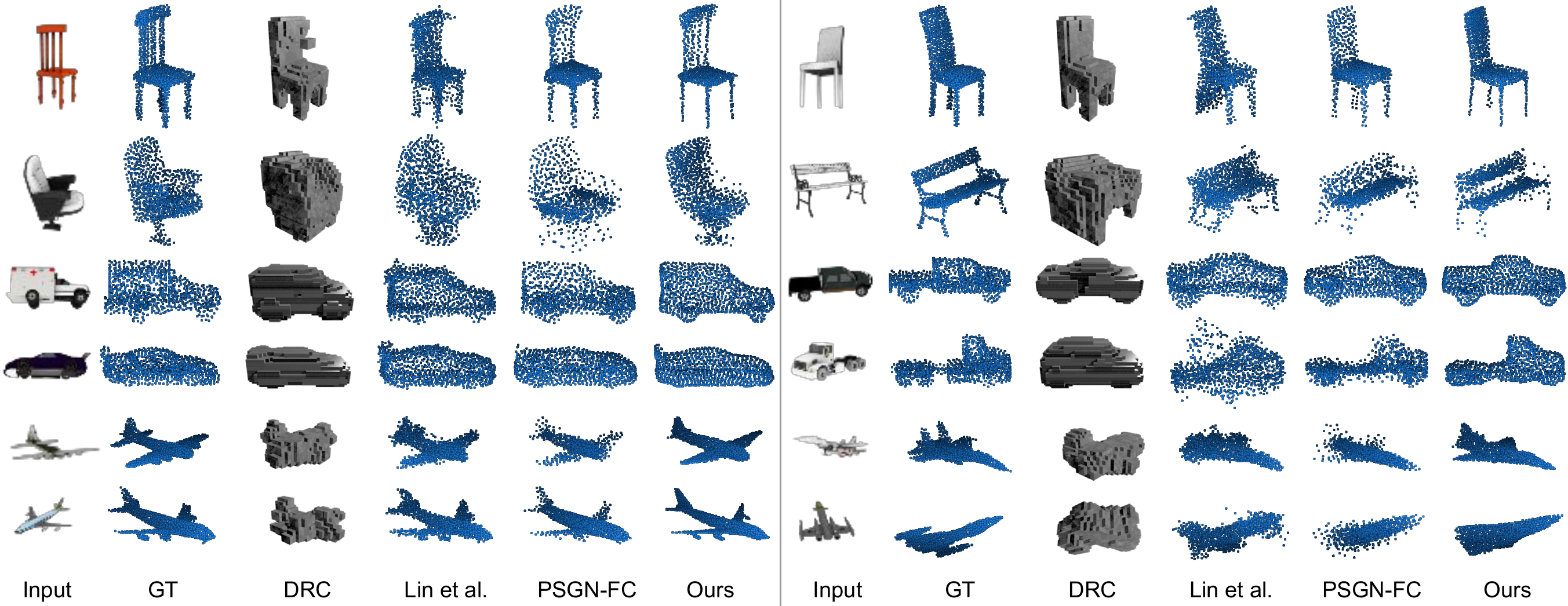}
\end{center}
\caption{Qualitative comparison on ShapeNet. Our network achieves better reconstructions, with high correspondence to the input image. While DRC outputs are blobby and lack concavity, Lin et al. predictions are rough and have a number of outlier points. In contrast, our predictions are sharper and more accurate. Our network also predicts finer structures like bars on the back of chairs, which the 3D-supervised network fails to capture.}
\label{fig:shapenet_mask}
\end{figure*}
\begin{figure*}[!htb]
\centering
\begin{center}
    \includegraphics[width=\linewidth]{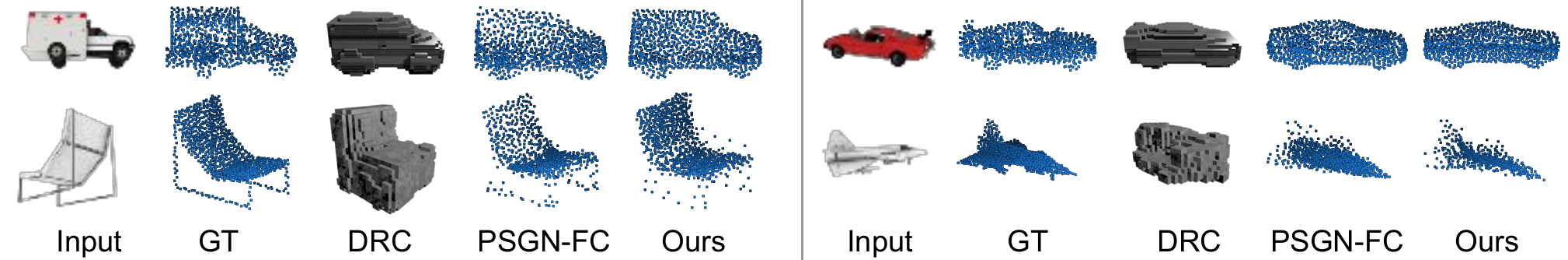}
\end{center}
\caption{Qualitative results for the multi-category experiment on ShapeNet.}
\label{fig:multi_cat}
\end{figure*}
\paragraph{Comparison} We benchmark our proposed approach against state-of-the-art 3D and 2D supervision works. For 3D supervision, we compare our work with the fully-connected decoder variant of ~\cite{fan2017point}, hereby referred to as PSGN-FC. The PSGN-FC network is trained with 3D point cloud supervision, and Chamfer distance is used as the loss function. We follow the same network architecture that is used for our 2D supervision experiments. For evaluating our approach against 2D supervision works,we compare with DRC~\cite{tulsiani2017multi}, which outputs voxelized 3D reconstructions. To evaluate DRC, we use the pre-trained models provided by the authors, and convert the representations to the point cloud format as described in the evaluation section. We also compare against Lin \textit{et al.}~\cite{lin2018learning}, who fuse depth maps from eight fixed views to obtain the point cloud. Since Lin \textit{et al.} predictions are dense, we apply farthest point sampling on the outputs to obtain $1024$ points for evaluation.
\paragraph{Results}


\begin{figure*}
    \centering
    \begin{minipage}[t]{.6\textwidth}
        \centering
        \includegraphics[width=\linewidth]{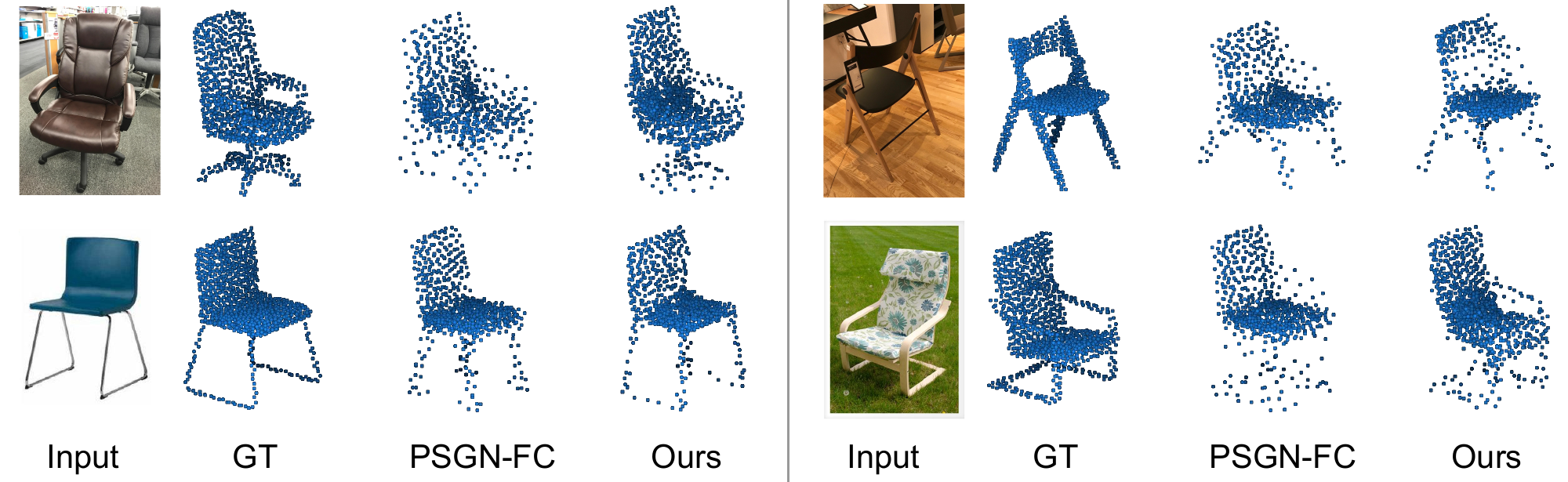}
        \caption{Qualitative comparison on the real-world Pix3D dataset~\cite{pix3d}. Our network fine-tuned with just a single mask as supervision, is able to effectively reconstruct from real world images. Shapes and finer details are better captured in comparison to the 3D supervised network trained only on ShapeNet.}
        \label{fig:pix3d_mask}
    \end{minipage}%
    \hspace{0.5em}
    \begin{minipage}[t]{.35\textwidth}
        \centering
        \includegraphics[width=\linewidth]{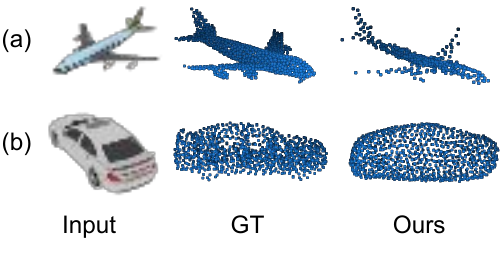}
        \caption{Failure modes. (a) Multi-category training produces narrower airplanes. (b) Cars trained with just a single mask as supervision have slightly deformed bodies.}
        \label{fig:failure_cases}
    \end{minipage}
\end{figure*}

\begin{figure*}[t]
\centering
\begin{center}
    \includegraphics[width=\linewidth]{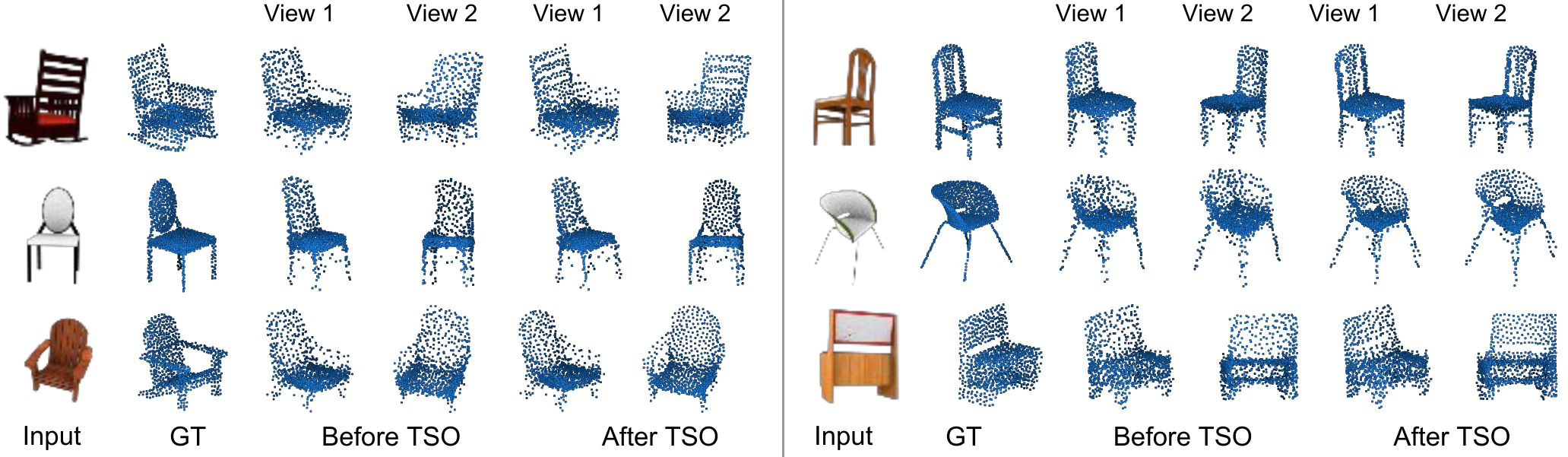}
\end{center}
\caption{Qualitative results showing the generated point clouds before and after test-stage optimization (TSO). The optimized point clouds display geometric and structural details (curved backrest, bars, \textit{etc}) present in the input image, that are absent in the initial predictions. Reconstruction results are shown from two different viewing angles so as to highlight the retention of structure when seen from a different view and the correspondence with the input image when seen from the same view.}
\label{fig:test_stage_optim}
\end{figure*}
\begin{table}[!htb]
\parbox{\linewidth}{
  \caption{Chamfer metrics on Pix3D (all values are scaled by 1000).
  (Fwd: GT\textrightarrow Pred, Bwd: Pred\textrightarrow GT. Chamfer: Fwd$+$Bwd.)}
  \centering
  \begin{tabular}{llll}
    \toprule
    Method          & Fwd    & Bwd      & Chamfer\\
    \midrule  
    PSGN-FC         & 5.04   & 5.06   & 10.1  \\
    Ours(joint)  & \textbf{4.44}      & \textbf{4.76}     & \textbf{9.2}   \\
    \bottomrule
  \end{tabular}
  \label{tab:pix3d_mask}
}
\end{table}
\begin{table}
\quad
\parbox{\linewidth}{
  \caption{Chamfer metrics on test stage optimization (metrics are scaled by 1000). (Fwd: GT\textrightarrow Pred, Bwd: Pred\textrightarrow GT. Chamfer: Fwd$+$Bwd.)}
  \centering
  \begin{tabular}{llll}
    \toprule
    Method          & Fwd    & Bwd & Chamfer\\
    \midrule  
    Pre-trained Net         & 2.45 &	2.19 & 4.64 \\
    TSO - update $E$, fix $D$ & 2.29 &	2.14 & 4.43\\
    TSO - update $E$ and $D$ & \textbf{2.28}	& \textbf{2.09} &\textbf{4.37}\\
    TSO - directly update $p$ & 2.36 &	2.13 & 4.49\\
    \bottomrule
  \end{tabular}
  \label{tab:test_stage}
}
\end{table}
\begin{figure*}
    \centering
    \begin{minipage}[t]{.25\textwidth}
        \centering
        \includegraphics[width=\linewidth]{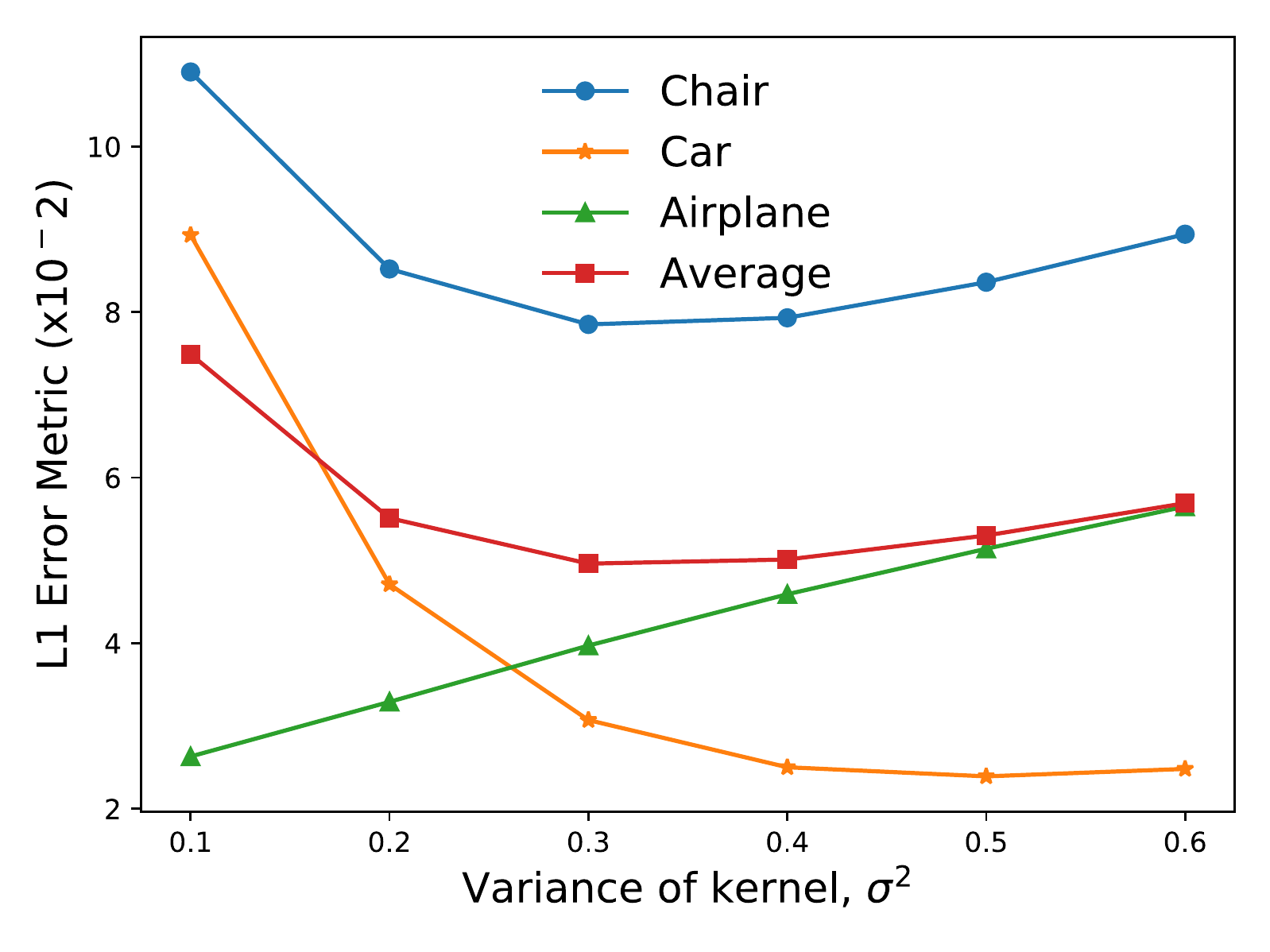}
        \scriptsize
        \caption{Variance of kernel, $\sigma^2$ vs $\mathcal{L}_1$ error for mask projections plotted for different categories.}
        \label{fig:variance_plot_mask}
    \end{minipage}%
    \hspace{0.5em}
    \begin{minipage}[t]{.7\textwidth}
        \centering
        \includegraphics[width=\linewidth]{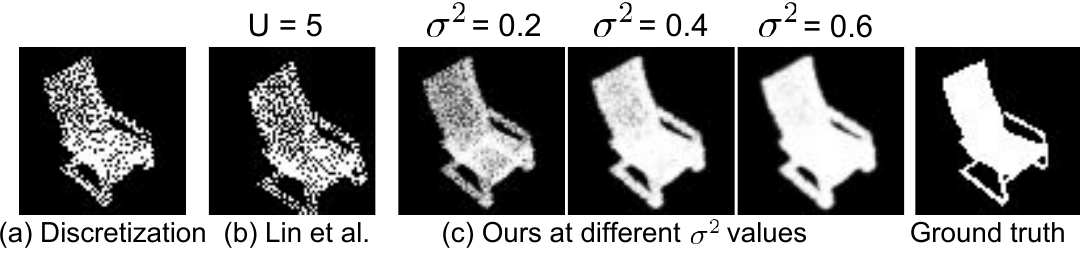}
        \scriptsize
        \caption{Projection maps for discretized projection, Lin \textit{et al.}, and our method. (a) and (b) have a lot of 'holes'. (c) Our projections are smooth due to the continuous approximation module. Here, $\sigma^2=0.4$ fills in holes while retaining finer details.}
        \label{fig:projection_maps}
    \end{minipage}
\end{figure*}
\begin{figure}
\centering
\begin{center}
    \includegraphics[width=0.65\linewidth]{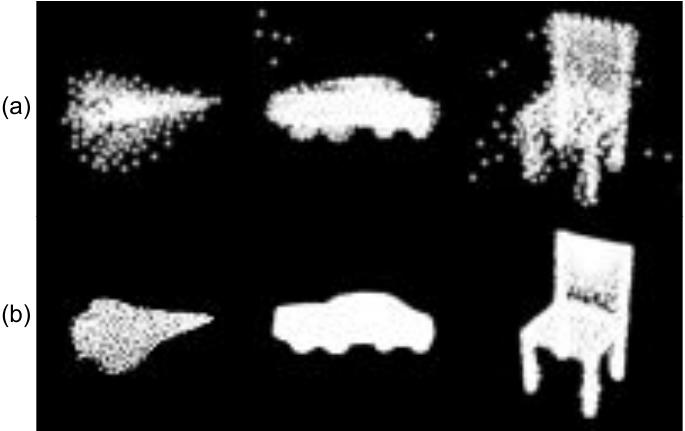}
\end{center}
\caption{Projected predictions trained (a) without and (b) with affinity loss, which helps in outlier point removal.}
\label{fig:affinity_loss}
\end{figure}
Table~\ref{tab:shapenet_mask} presents the results on the ShapeNet dataset with comparisons against the 3D-supervised PSGN-FC, 2D-supervised DRC and depth-fusion-based Lin et al. We significantly outperform DRC while achieving results comparable to the 3D supervised PSGN-FC. It is interesting to note that our approach, with just foreground masks as supervision, outperforms even the depth-based approaches of DRC and Lin et al. Fig.~\ref{fig:shapenet_mask} shows qualitative results on ShapeNet. We observe that unlike DRC, our approach accurately reconstructs structures with concavities. Lin et al. predictions are rough and have a number of outlier points, while our predictions are sharper and more accurate. Our network is also able to better predict finer details, \textit{e.g.} hollow regions in the back of chairs, which the 3D-supervised method fails to capture. We also train a single network on all three categories. We observe that the network performs comparably to the single category variants (Table~\ref{tab:shapenet_mask} and Fig.\ref{fig:multi_cat}).

\subsection{3D Reconstruction on Pix3D}

To show the efficacy of our approach on real data, we consider Pix3D, a real world dataset with 3D CAD models and corresponding 2D images from multiple view-points. We use the  3D models only for evaluation. We randomly select $80\%$ of the dataset as the train set and the rest as the test set. We show that fine-tuning a PSGN-FC network pre-trained on ShapeNet with the additional 2D training data results in improved performance. To adapt to the domain of real world images, the PSGN-FC network is trained with synthetic ShapeNet images overlaid on random natural backgrounds, as done in~\cite{tulsiani2017multi}. Results are reported in Table~\ref{tab:pix3d_mask}. Qualitative
results are presented in Fig.~\ref{fig:pix3d_mask}. We observe that as in the case of the synthetic
dataset, our approach results in more faithful reconstructions. The finer details present in the leg and 
handle regions of chairs are effectively captured. This 
demonstrates the applicability of our training methodology for real world scenarios.
\subsection{Test Stage optimization}

To evaluate our test stage optimization (TSO) approach, we train a PSGN-FC network to 
reconstruct point clouds aligned with the input image view. We then perform the 
optimization step at the test stage for every input image. We set the distance function $d$ in Eqn.~\ref{eqn:tso} to be the Chamfer distance between the initial and optimized point clouds in all our experiments. Table~\ref{tab:test_stage} shows quantitative metrics for the TSO variant where both the encoder and decoder are updated by optimizing the projection and regularization losses (Eqn.~\ref{eqn:tso}). We report the Chamfer distance along with the corresponding forward and backward losses on the test set. We observe that all three variants of TSO lead to improved performance compared to the intial prediction from the pre-trained network. In our qualitative study, we observed that the TSO variant with the decoder fixed had limited capacity to capture the finer details present in the image, while the other two variants (($E$,$D$) updated, and only $p$ updated) performed better in generating point clouds that correspond to the input image. Fig.~\ref{fig:test_stage_optim} shows qualitative results on samples from the test set before and after running the optimization routine. We observe that the reconstructions match the input image from the visible view, while preserving the structure in the occluded regions. The optimized point clouds display geometric and structural details (curved backrest, bars, \textit{etc}) present in the input image, that are absent in the initial predictions. 

\section{Discussion}
\paragraph{Variance of kernel} We plot the category-averaged $\mathcal{L}_1$ error between the ground truth mask and corresponding projection for different $\sigma^2$ values (Eqn.~\ref{eq:kernel}) in Fig.~\ref{fig:variance_plot_mask}. Projections for a sample model are shown in Fig.~\ref{fig:projection_maps}c. Lower $\sigma^2$ values result in holes whereas higher values fill large areas. 
$\sigma^2$ is set to $0.4$ for chair and car, and $0.1$ for airplane. In the multi-category setup, it is set to 0.4 for optimal performance. 
For comparison, we also display the obtained projection map from a naive discretization-based method (Fig.~\ref{fig:projection_maps}a) and Lin \textit{et al.}'s method (with upsampling factor $U=5$) (Fig.~\ref{fig:projection_maps}b).
\paragraph{Effect of number of projections} We study the effect of number of ground 
truth projections per image used during training (Table~\ref{tab:n_views}). We observe 
that the network is able to reconstruct well with just a single mask as supervision. 
The performance improves with two masks and stays constant with higher number of 
projections. 
\paragraph{Role of Affinity Loss} To analyze the role of the affinity loss $\mathcal{L}_{\mathrm{aff}}$ (Eqn.~\ref{eq:loss_aff}), we train a network with only cross entropy loss $\mathcal{L}_{bce}$ (Eqn.~\ref{eq:loss_bce}). We observe that the reconstructions are noisy with a lot of outlier points, resulting in higher Chamfer scores. The addition of $\mathcal{L}_{\mathrm{aff}}$, results in better reconstructions (Fig.~\ref{fig:affinity_loss}). However, in the case of chairs, we notice that a high weightage to $\mathcal{L}_{\mathrm{aff}}$ delays the emergence of thinner structures like legs. Hence, we reduce the the weight of $\mathcal{L}_{\mathrm{aff}}$ after a fixed number of iterations. The loss weighing strategy  between $\mathcal{L}_{bce}$ and $\mathcal{L}_{\mathrm{aff}}$ is provided in the supplementary.

\paragraph{Failure modes} 
We analyze the failure modes for our method in Fig.~\ref{fig:failure_cases}.
\begin{enumerate*}[label=\textbf{(\alph*)}]
    \item Certain instances of airplanes have a narrower body in comparison to the ground truth for multi-category trained models. Choosing $\sigma^2$ values per category alleviates this effect.
    \item Cars trained with only single view mask supervision (V=1) have a slight deformation in the depth dimension. Using an additional view during supervision corrects this.
\end{enumerate*}
\begin{table}
  \setlength{\tabcolsep}{4pt}
  \caption{Dependency of reconstruction performance on number of views. Chamfer scores are scaled by 1000}
  \centering
  \begin{tabular}{lllllllll}
    \toprule
    Num Views       & 1     & 2     & 3     & 4     & 5     & 6     & 8 \\
    \midrule  
    Airplane        & 2.40  & 2.02  & 2.0   & 2.0   & 1.98  & 1.99  & 2.01 \\
    Car             & 3.47  & 1.68  & 1.65  & 1.65  & 1.68  & 1.65  & 1.68 \\
    Chair           & 4.53  & 4.41  & 4.35  & 4.36  & 4.43  & 4.42  & 4.56 \\
    \bottomrule
  \end{tabular}
  \label{tab:n_views}
\end{table}
\section{Conclusion}
We introduced a continuous approximation projection module for a differentiable and accurate rendering of 3D point clouds, to enable weakly supervised 3D object reconstruction. A loss formulation was introduced in the training procedure to improve reconstructions. 
We highlighted the utility of such an approach in real world scenarios by improving the reconstruction performance using as little as a single mask as supervision. Quantitative and qualitative evaluation on synthetic and real-world datasets show that the generated point clouds are of better quality in comparison to the current state-of-art projection-based reconstruction methods. Furthermore, we also demonstrated that the presence of object masks at test stage can be utilized to obtain highly corresponding 3D reconstructions. In the future, we would like to explore ways of extending the projection framework to obtain depth maps, color images and any other features associated with the object.

\section{Acknowledgement}
This work was funded by the Robert Bosch Centre for Cyber Physical Systems, IISc (Project ID: RBCO0052).

\bibliography{egbib}
\bibliographystyle{aaai}
\end{document}


\title{CAPNet: Continuous Approximation Projection For 3D Point Cloud Reconstruction Using 2D Supervision: Supplementary Material}
\author{Navaneet K L\thanks{equal contribution},
Priyanka Mandikal\printfnsymbol{1},
Mayank Agarwal, \and
R. Venkatesh Babu \\
Video Analytics Lab, CDS, Indian Institute of Science, Bangalore, India \\
navaneetl@iisc.ac.in, priyanka.mandikal@gmail.com, mayankgrwl97@gmail.com, venky@iisc.ac.in
}
\maketitle

\section{Network Architecture}


\begin{table}[h]
\centering
\begin{tabular}{|c|c|c|c|c|}
\hline
S.No. & Layer   & \begin{tabular}[c]{@{}c@{}}Filter Size/\\ Stride\end{tabular} & Output Size  \\ \hline\hline
1     & conv   & 3x3/1                                                         & 64x64x32   \\
2     & conv   & 3x3/1                                                         & 64x64x32   \\
3     & conv   & 3x3/2                                                         & 32x32x64     \\
4     & conv   & 3x3/1                                                         & 32x32x64     \\
5     & conv   & 3x3/1                                                         & 32x32x64     \\
6     & conv   & 3x3/2                                                         & 16x16x128    \\
7     & conv   & 3x3/1                                                         & 16x16x128    \\
8     & conv   & 3x3/1                                                         & 16x16x128    \\
9     & conv   & 3x3/2                                                         & 8x8x256    \\
10    & conv   & 3x3/1                                                         & 8x8x256    \\
11    & conv   & 3x3/1                                                         & 8x8x256    \\
16    & conv   & 5x5/2                                                         & 4x4x512      \\
17    & linear & -                                                             & 128          \\ \hline
\end{tabular}
\caption{Image Encoder Architecture}
\label{tab:image_encoder_architecture}
\end{table}

\begin{table}[h]
\centering
\begin{tabular}{|c|c|c|}
\hline
S.No. & Layer   & Output Size\\ \hline\hline
1     & linear & 256         \\
2     & linear & 256         \\
3     & linear & 1024*3      \\ \hline
\end{tabular}
\caption{Decoder Architecture}
\label{tab:point_cloud_decoder_architecture}
\end{table}
\section{Loss Weighing Strategy}
\begin{table}[!h]
\centering
\begin{tabular}{cccc}
\toprule
$\lambda$=0.5    & $\lambda$=1  & $\lambda$=5 \\
\midrule  
4.63      & \textbf{4.42}      & 5.3 \\
\bottomrule
\end{tabular}
\caption{Chamfer metric for models trained with different weightage $\lambda$ (scaled by $1e^{4}$) for $\mathcal{L}_{\textrm{aff}}$ (Eqn.5, main paper). Based on this, we set $\lambda$ to be 1 in all our experiments. During training, $\lambda$ is reduced to $0.02$ of its initial value upon loss saturation.}
\label{tab:loss_weighing}
\end{table}
\begin{figure*}[t]
\centering
\begin{center}
    \includegraphics[width=\linewidth]{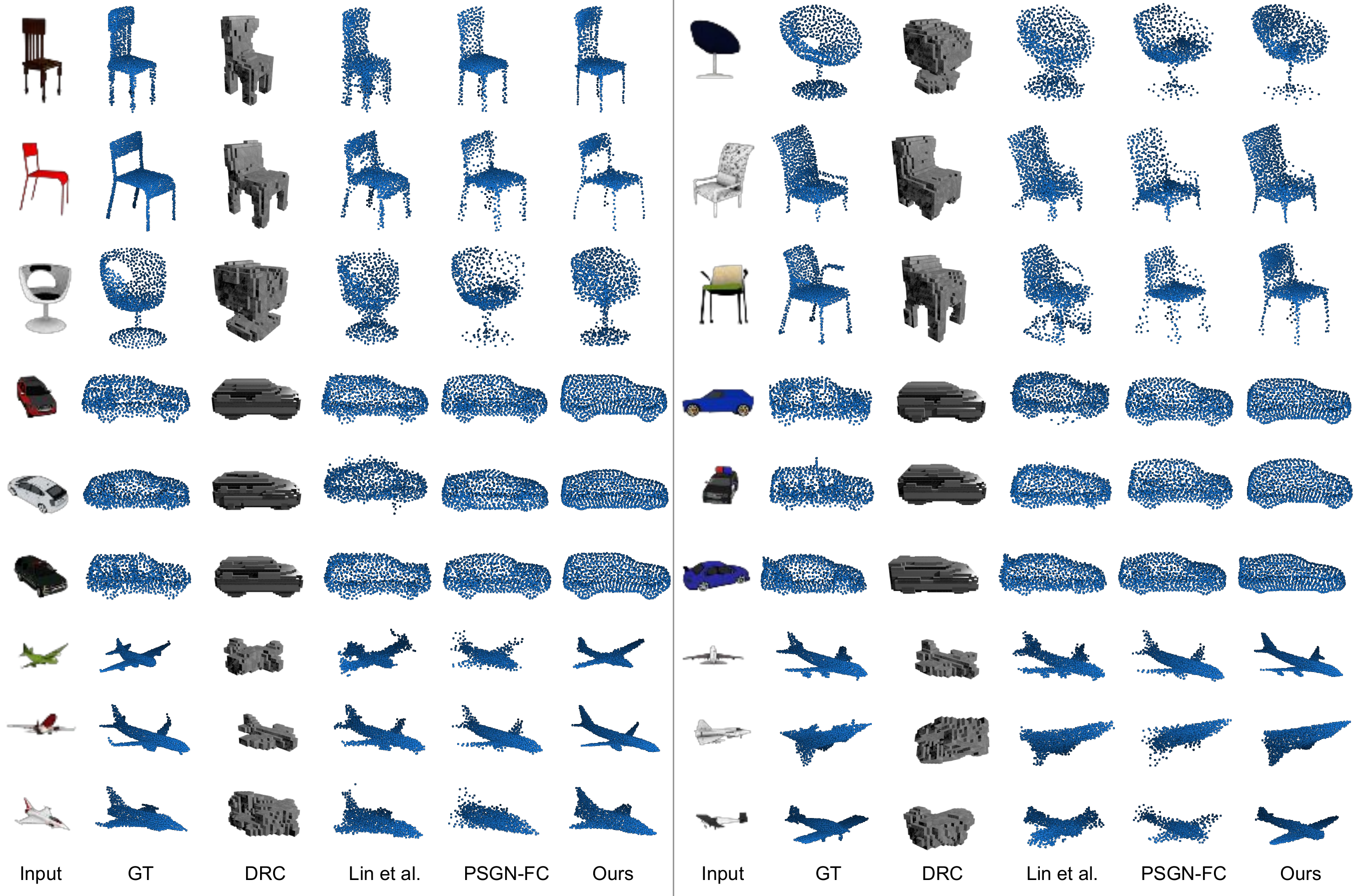}
\end{center}
\caption{Qualitative comparison on ShapeNet dataset.}
\label{fig:supp_shapenet_mask}
\end{figure*}
\begin{figure*}[t]
\centering
\begin{center}
    \includegraphics[width=\linewidth]{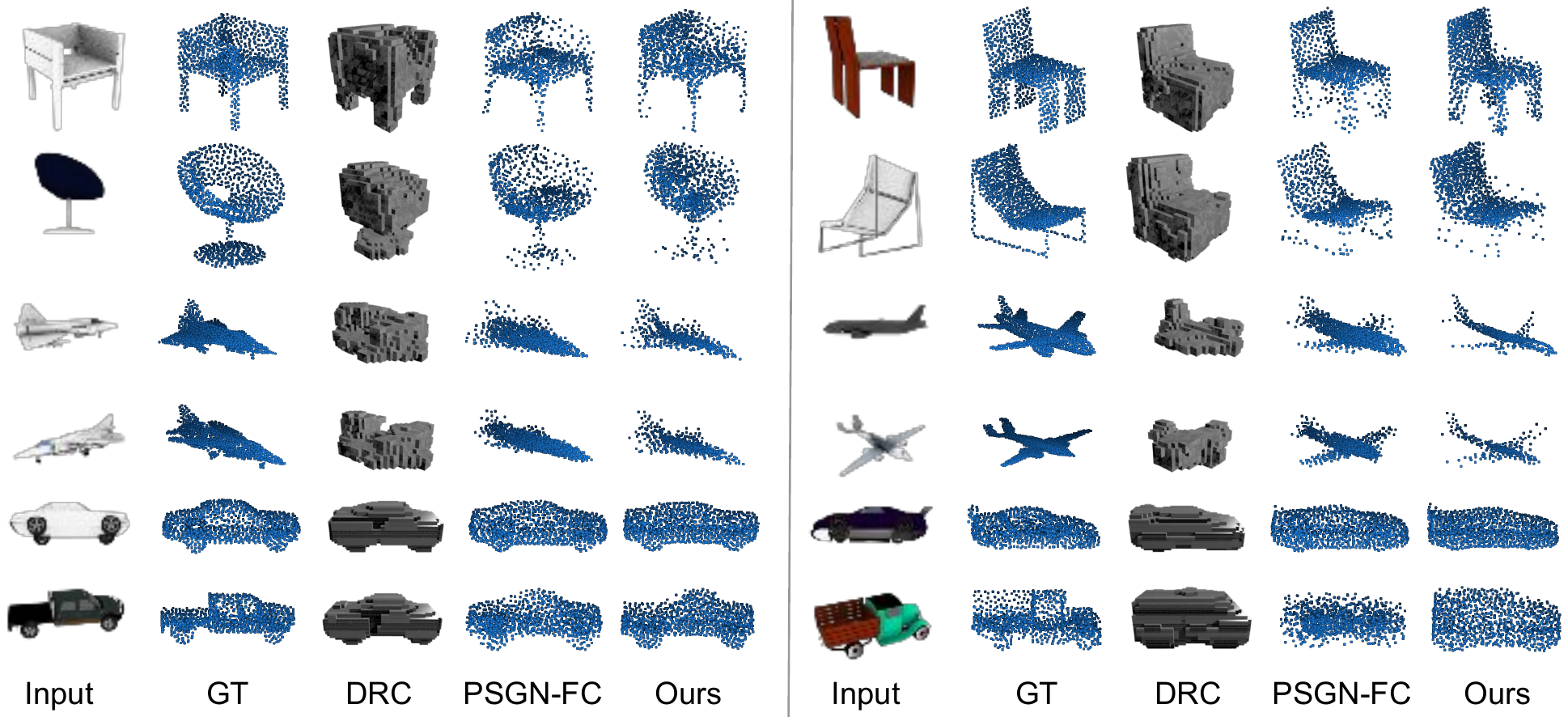}
\end{center}
\caption{Qualitative comparison for multi-category network on ShapeNet dataset.}
\label{fig:supp_multi_cat}
\end{figure*}
\section{3D Reconstruction Results on ShapeNet and Pix3D Datasets}
\begin{figure*}[t]
\centering
\begin{center}
    \includegraphics[width=\linewidth]{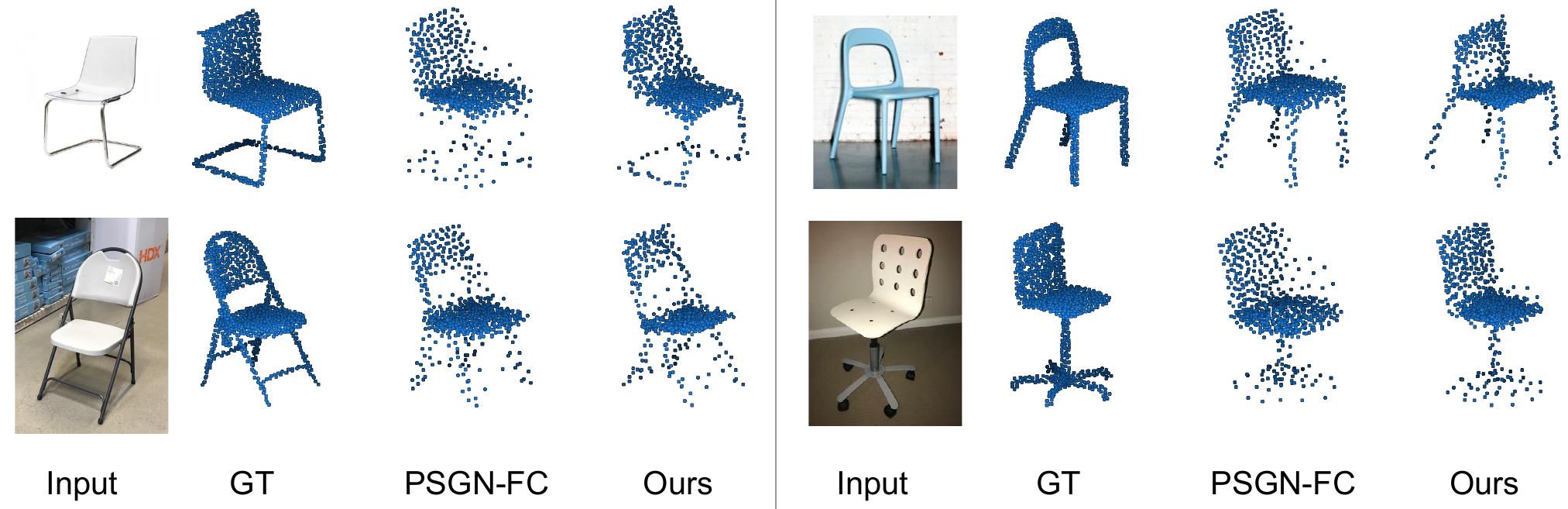}
\end{center}
\caption{Qualitative comparison on Pix3D dataset.}
\label{fig:supp_pix3d_mask}
\end{figure*}

Qualitative results on ShapeNet dataset are shown in Fig.~\ref{fig:supp_shapenet_mask}. Mutli-category experiment results are shown in Fig.~\ref{fig:supp_multi_cat}. Fig.~\ref{fig:supp_pix3d_mask} presents the corresponding results on the real world pix3D dataset. 
\section{Test Stage Optimization}
\begin{figure*}[t]
\centering
\begin{center}
    \includegraphics[width=\linewidth]{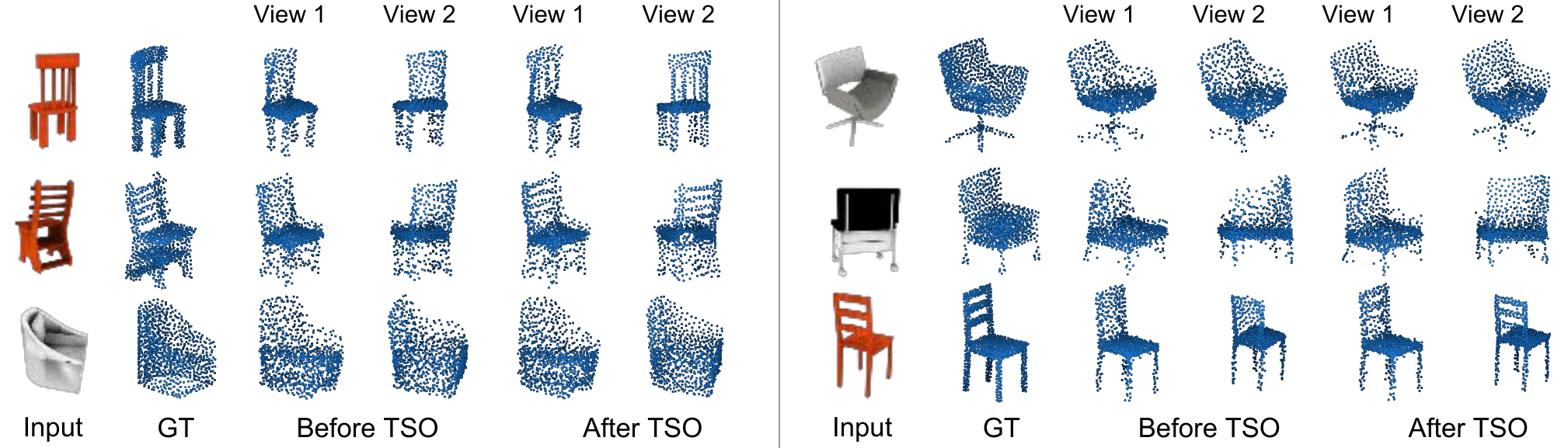}
\end{center}
\caption{Qualitative results for test stage optimization.}
\label{fig:supp_test_optim}
\end{figure*}

Qualitative results for test stage optimization is presented in Fig.~\ref{fig:supp_test_optim}.
